\documentclass[journal]{IEEEtran}
\usepackage{cite}
\usepackage{amsmath,amssymb,amsfonts}
\usepackage{algorithmic}
\usepackage{graphicx}
\usepackage{algorithm,algorithmic}
\usepackage{hyperref}
\hypersetup{hidelinks=true}
\usepackage{textcomp}
\usepackage{booktabs}
\usepackage{multirow}
\usepackage{xcolor}
\usepackage{subcaption}

\def\BibTeX{{\rm B\kern-.05em{\sc i\kern-.025em b}\kern-.08em
    T\kern-.1667em\lower.7ex\hbox{E}\kern-.125emX}}
\begin{document}
\title{Curriculum-Driven 3D CT Report Generation via Language-Free Visual Grafting and Zone-Constrained Compression}

\author{V.~K.~Cody~Bumgardner,
        Mitchell~A.~Klusty,
        Mahmut~S.~Gokmen,
        and~Evan~W.~Damron \\
        \textit{Center for Applied Artificial Intelligence, University of Kentucky, Lexington, KY 40506 USA} \\
        \{cody, mitchell.klusty, m.gokmen, Evan.Damron\}@uky.edu}

\maketitle

\begin{abstract}
Automated radiology report generation from three-dimensional computed tomography (CT) volumes remains a formidable challenge due to the extreme sequence lengths of volumetric data, severe class imbalance between normal and pathological findings, and the tendency of large language models (LLMs) to ignore grafted visual tokens in favor of linguistic priors. We present Ker-VLJEPA-3B, a four-phase curriculum learning framework for free-text report generation from thoracic CT volumes. A frozen, self-supervised visual encoder produces representations unconstrained by text labels or language model objectives, and a phased training curriculum progressively adapts a Llama~3.2~3B decoder to ground its output in these visual features. The visual backbone (LeJEPA ViT-Large) is trained entirely via self-supervised joint-embedding prediction on unlabeled CT volumes, with no text supervision of any kind. Unlike contrastive vision-language encoders (CLIP, BiomedCLIP) that entangle visual representations with linguistic biases from training, our language-free backbone produces modality-pure representations optimized solely for visual understanding. All vision-language alignment is deferred to the curriculum's bridge and generation phases, enabling principled control over when and how modalities interact. This decoupled design is modality-agnostic: the same framework can integrate any self-supervised encoder, whether trained on imaging, genomic, or sensor data, into a language model without requiring paired text during foundation model training. Our approach further introduces several methodological innovations: (1)~zone-constrained cross-attention that compresses variable-length slice embeddings into 32 spatially-grounded visual tokens preserving anatomical localization; (2)~PCA whitening of anisotropic LLM embeddings enabling effective contrastive alignment; (3)~a positive-findings-only training strategy that eliminates posterior collapse caused by normal-text gradient domination; (4)~a warm bridge initialization technique that transfers converged vision-to-LLM projection weights across curriculum phases; and (5)~selective cross-attention freezing with elastic weight consolidation for catastrophic forgetting prevention during narrative fine-tuning. Evaluated on the CT-RATE benchmark (2{,}984 validation volumes, 18 thoracic abnormality classes) using the official RadBERT label extraction protocol, Ker-VLJEPA-3B achieves a macro~F1 of 0.429, surpassing the previous state-of-the-art (U-VLM, macro~F1\,=\,0.414) by 3.6\%, with per-class threshold optimization yielding macro~F1\,=\,0.448 (+8.2\%). A comprehensive visual token ablation study confirms that 56.6\% of generation quality derives from patient-specific visual content. Code and model weights are publicly available.\footnote{\url{https://huggingface.co/IBI-CAAI/Ker-VLJEPA-3B}}
\end{abstract}

\begin{IEEEkeywords}
Computed tomography, medical report generation, multimodal learning, vision-language models, embedding grafting, cross-attention, curriculum learning.
\end{IEEEkeywords}

\section{Introduction}
\label{sec:introduction}

Thoracic computed tomography (CT) is a cornerstone of modern diagnostic radiology, generating volumetric datasets that encode rich three-dimensional anatomical and pathological information across hundreds of axial slices. The interpretation of these volumes and the production of structured narrative reports constitutes one of the most labor-intensive tasks in clinical radiology, contributing to radiologist burnout and diagnostic variability~\cite{ctrate}. Automated report generation from 3D CT data has the potential to streamline clinical workflows, improve consistency, and serve as a decision-support tool for radiologists.

Recent advances in vision-language models (VLMs) have demonstrated impressive capabilities in natural image captioning and two-dimensional chest X-ray interpretation~\cite{flamingo,clip}. However, extending these approaches to 3D thoracic CT introduces fundamental challenges that remain largely unsolved. First, volumetric imaging generates massive data structures where a single scan may contain 300--600 axial slices, each encoded as a high-dimensional feature vector, creating sequence lengths that far exceed the context windows of standard large language models (LLMs). Second, critical pathologies such as small lung nodules or early interstitial changes may occupy less than 1\% of the total voxel space, creating extreme class imbalance that biases models toward generating safe, generic normal-anatomy descriptions. Third, when visual tokens are grafted into a pre-trained LLM's embedding stream, the overwhelming strength of linguistic priors causes the model to attend primarily to textual context while largely ignoring the injected visual information, a phenomenon we term \textit{posterior collapse} in the generative setting.

Existing approaches to 3D CT report generation have achieved varying degrees of success on the CT-RATE benchmark~\cite{ctrate}. CT-CLIP~\cite{ctclip} demonstrated zero-shot classification via contrastive pre-training (macro~F1\,=\,0.194). CT-CHAT~\cite{ctchat} fine-tuned a vision-language model achieving macro~F1\,=\,0.287. BTB3D~\cite{btb3d} employed a custom 3D encoder (macro~F1\,=\,0.354). Most recently, U-VLM~\cite{uvlm} established the state-of-the-art at macro~F1\,=\,0.414, introducing two key innovations: progressive training from segmentation to classification to report generation, and multi-layer visual injection that routes hierarchical U-Net encoder features to corresponding language model layers. Our work builds upon U-VLM's insight that multi-layer injection improves report quality, extending it with Flamingo-style gated cross-attention~\cite{flamingo} at intermediate LLM layers and continuous visual grounding during autoregressive decoding. However, none of the existing methods explicitly address the posterior collapse problem that arises from the severe gradient imbalance between normal and pathological text tokens during generative training.

A further limitation shared by all prior approaches is their reliance on vision encoders trained with text supervision. CT-CLIP and CT-CHAT use contrastive image-text pre-training; U-VLM uses a segmentation-pretrained U-Net whose training labels are themselves derived from clinical annotations. In all cases, linguistic or semantic biases are baked into the visual representations before generation training begins. This entanglement means the encoder may under-represent visually salient features that are rarely described in text, and the resulting representations are tightly coupled to the specific language task for which they were trained.

In this work, we present \textbf{Ker-VLJEPA-3B}, a Vision-Language extension of the KerJEPA~\cite{kerjepa} family of kernel-regularized Joint-Embedding Predictive Architectures~\cite{ijepa}, realized as a four-phase curriculum learning framework that systematically addresses these challenges through a series of methodological innovations. A central design principle is the \textit{complete decoupling of visual representation learning from language}: the visual backbone (LeJEPA ViT-Large~\cite{lejepa_paper,lejepa}) is trained entirely via self-supervised joint-embedding prediction on unlabeled CT volumes, with no text, labels, or linguistic signal of any kind. This produces modality-pure visual representations optimized solely for capturing anatomical and pathological structure. All vision-language alignment is deferred to the curriculum's bridge and generation phases, where it can be controlled precisely. Because the framework imposes no assumptions about the modality of the input encoder, the same curriculum and bridge architecture could integrate any self-supervised foundation model, whether trained on imaging, genomic sequences, or time-series sensor data, into a language model without requiring paired text during foundation model training. Our key contributions are:

\begin{itemize}
    \item \textbf{Language-free, modality-agnostic architecture}: we demonstrate that a visual encoder trained with purely self-supervised objectives (no text, no labels) can be effectively grafted into a pre-trained LLM via a curriculum bridge, achieving state-of-the-art report generation without any text supervision in the foundation model. This decoupled design generalizes beyond imaging to any modality for which a self-supervised encoder exists.

    \item \textbf{Zone-constrained cross-attention} for volumetric compression: a spatial attention mechanism that compresses variable-length CT slice embeddings into 32 fixed-size visual tokens, where each token attends exclusively to slices within its anatomical zone along the z-axis, preserving spatial localization of pathology.

    \item \textbf{PCA whitening for contrastive alignment}: we identify and resolve the catastrophic anisotropy of LLM text embeddings (mean pairwise cosine similarity\,=\,0.949) that renders standard contrastive learning ineffective, achieving an 11.8$\times$ improvement in discriminability through projection to an isotropic 256-dimensional space.

    \item \textbf{Positive-findings-only training}: a training data reformulation that eliminates the 90\% normal-text gradient domination causing posterior collapse, enabling sustained generative performance across 15+ training epochs where all prior approaches collapsed within 1--4 epochs.

    \item \textbf{Warm bridge initialization}: a technique for transferring converged vision-to-LLM projection and cross-attention weights across curriculum phases, providing immediate convergence (epoch~1 F1\,=\,0.425 vs.\ 0.360 cold start) and enabling the model to surpass rather than merely recover to prior-phase performance.

    \item \textbf{Selective cross-attention freezing with EWC}: a principled approach to narrative fine-tuning that decouples visual grounding (cross-attention) from linguistic style (LoRA), preserving pathology detection while adapting to authentic radiologist prose.

    \item \textbf{Comprehensive ablation study} demonstrating that 56.6\% of generation quality derives from patient-specific visual token content, with semantic binding analysis showing 2$\times$ stronger visual contribution on pathology-specific words.
\end{itemize}

Ker-VLJEPA-3B achieves macro~F1\,=\,0.429 on the CT-RATE benchmark (2{,}984 validation volumes), surpassing U-VLM by +3.6\%, establishing a new state-of-the-art for automated 3D CT report generation.

\section{Related Work}
\label{sec:related}

\subsection{Vision-Language Models in Medical Imaging}

The integration of visual and linguistic modalities for medical image analysis has progressed rapidly from contrastive pre-training approaches~\cite{clip} to generative vision-language models~\cite{flamingo}. In the medical domain, BiomedCLIP and related models demonstrated the value of domain-specific contrastive pre-training for 2D radiograph understanding. For chest X-ray report generation, models such as CheXpert and R2Gen established effective encoder-decoder pipelines. However, these 2D approaches fundamentally cannot capture the volumetric spatial relationships critical for CT interpretation.

\subsection{3D CT Understanding and Report Generation}

The CT-RATE dataset~\cite{ctrate} established the first large-scale benchmark for 3D CT report generation, comprising 50{,}188 thoracic CT volumes with 18 binary abnormality labels and radiologist-authored narrative reports. CT-CLIP~\cite{ctclip} adapted contrastive learning to 3D volumes via a ViT-based encoder, achieving zero-shot classification but not text generation. CT-CHAT~\cite{ctchat} extended this with a chat-style VLM interface but showed limited clinical accuracy (macro~F1\,=\,0.287). BTB3D~\cite{btb3d} introduced 3D Haar wavelet compression with causal convolutions to produce compact frequency-aware tokens (macro~F1\,=\,0.354). U-VLM~\cite{uvlm} proposed hierarchical vision-language modeling with a segmentation-pretrained U-Net encoder and multi-layer visual injection that routes encoder features at different scales to corresponding LLM layers, achieving macro~F1\,=\,0.414. Our multi-layer injection strategy differs from U-VLM's in two respects: we employ Flamingo-style gated cross-attention adapters~\cite{flamingo} rather than direct feature routing, and our cross-attention hooks fire on every autoregressive decode step, providing continuous visual grounding during generation rather than only at prefill.

Beyond the CT-RATE benchmark, the broader 3D medical VLM field has developed several paradigms for handling volumetric data. Med3DVLM~\cite{med3dvlm} employs decomposed 3D convolutions (DCFormer) coupled with a dual-stream MLP-mixer projector that blends low-level spatial features with high-level semantic representations. M3D-LaMed~\cite{m3dlamed} uses a 3D spatial pooling perceiver that reconstructs visual tokens into 3D coordinates before aggressive cross-attention compression. Med-2E3~\cite{med2e3} introduces text-guided inter-slice scoring, where a dot-product attention mechanism dynamically weights slice relevance conditioned on the clinical query. SCALE-VLP~\cite{scalevlp} proposes soft-weighted contrastive alignment that replaces binary matching with continuous, semantics-aware distances. RadZero3D~\cite{radzero3d} adapts the V-JEPA~2 video foundation model by treating CT depth as a temporal sequence.

A critical observation is that \textit{all} of these methods employ vision encoders trained with some form of text supervision, including contrastive image-text pre-training (CT-CLIP, Med3DVLM, SCALE-VLP), segmentation labels derived from clinical annotations (U-VLM), or text-conditioned adaptation (RadZero3D). Furthermore, most require either specialized 3D encoders with cubic computational scaling~\cite{med3dvlm}, aggressive slice filtering to reduce the input to a manageable size~\cite{med2e3}, or fixed-resolution volume resizing that discards native spatial information. Our approach fundamentally differs on both axes: the visual backbone is trained with \textit{no text supervision whatsoever}, and all slices are processed without filtering or resizing via zone-constrained cross-attention that compresses variable-length sequences into exactly 32 spatially-grounded tokens.

\subsection{Embedding Grafting and Vision-Language JEPAs}

The technique of grafting non-textual embeddings into the latent space of decoder-only transformers was pioneered by Flamingo~\cite{flamingo}, which introduced gated cross-attention layers interleaved with frozen LLM blocks. Subsequent work in LLaVA, MiniGPT-4, and Qwen-VL demonstrated that visual embeddings can be effectively projected into the LLM's token embedding space. Yue~{\it et al.}~\cite{grafting} formalized the zero-shot grafting problem and showed that LLM surrogates can bridge vision encoders to language decoders without paired training data, highlighting the importance of manifold alignment in the grafting process. Recently, VL-JEPA~\cite{vljepa} extended the JEPA paradigm to vision-language modeling by predicting continuous text embeddings rather than autoregressive tokens, achieving strong performance on video understanding with 50\% fewer trainable parameters. Concurrently, LLM-JEPA~\cite{llmjepa} demonstrated that embedding-space training objectives can outperform standard token-space reconstruction for LLM fine-tuning, providing theoretical motivation for our use of a JEPA embedding prediction loss as a non-autoregressive semantic anchor alongside the language modeling objective. While VL-JEPA operates on natural video with a discriminative embedding objective and LLM-JEPA targets unimodal language tasks, our work extends the JEPA framework in a complementary direction: we combine kernel-regularized embedding prediction~\cite{kerjepa} with autoregressive text generation for the multimodal medical domain, where free-text narrative output from visual input is required. Furthermore, adapting embedding grafting to 3D medical volumes, where the visual token count must be carefully controlled to avoid overwhelming the LLM's context window, requires specialized compression mechanisms and multi-phase training strategies not addressed by existing frameworks.

\subsection{Curriculum Learning and Continual Learning}

Curriculum learning~\cite{curriculum} has shown consistent benefits across machine learning domains by organizing training from simpler to more complex tasks. In multimodal settings, phased training has been used to first align representations and then train generative capabilities. Our four-phase curriculum extends this principle with innovations in cross-phase weight transfer (warm bridge) and selective parameter freezing. Elastic Weight Consolidation (EWC)~\cite{ewc} provides a principled mechanism for continual learning by adding a quadratic penalty that discourages important parameters from deviating from previously learned values, which we employ to prevent catastrophic forgetting during narrative fine-tuning.

\section{Methods}
\label{sec:methods}

\subsection{Problem Formulation}

Given a thoracic CT volume, we first extract per-slice visual features using a pre-trained Guided-Chest-CT-LeJEPA~\cite{lejepa_paper,lejepa} backbone, yielding a sequence of slice embeddings $\mathbf{S} = \{\mathbf{s}_1, \mathbf{s}_2, \ldots, \mathbf{s}_N\} \in \mathbb{R}^{N \times d_v}$ where $N \leq 600$ and $d_v = 1024$. The objective is to generate a free-text narrative radiology report $\mathbf{y} = (y_1, y_2, \ldots, y_T)$ that accurately describes the thoracic findings present in the scan. Clinical accuracy is evaluated by extracting 18 binary abnormality labels from the generated text using a RadBERT classifier and computing macro-averaged F1 against ground-truth labels.

\subsection{Architecture Overview}

\begin{figure*}[t]
    \centering
    \includegraphics[width=0.95\textwidth]{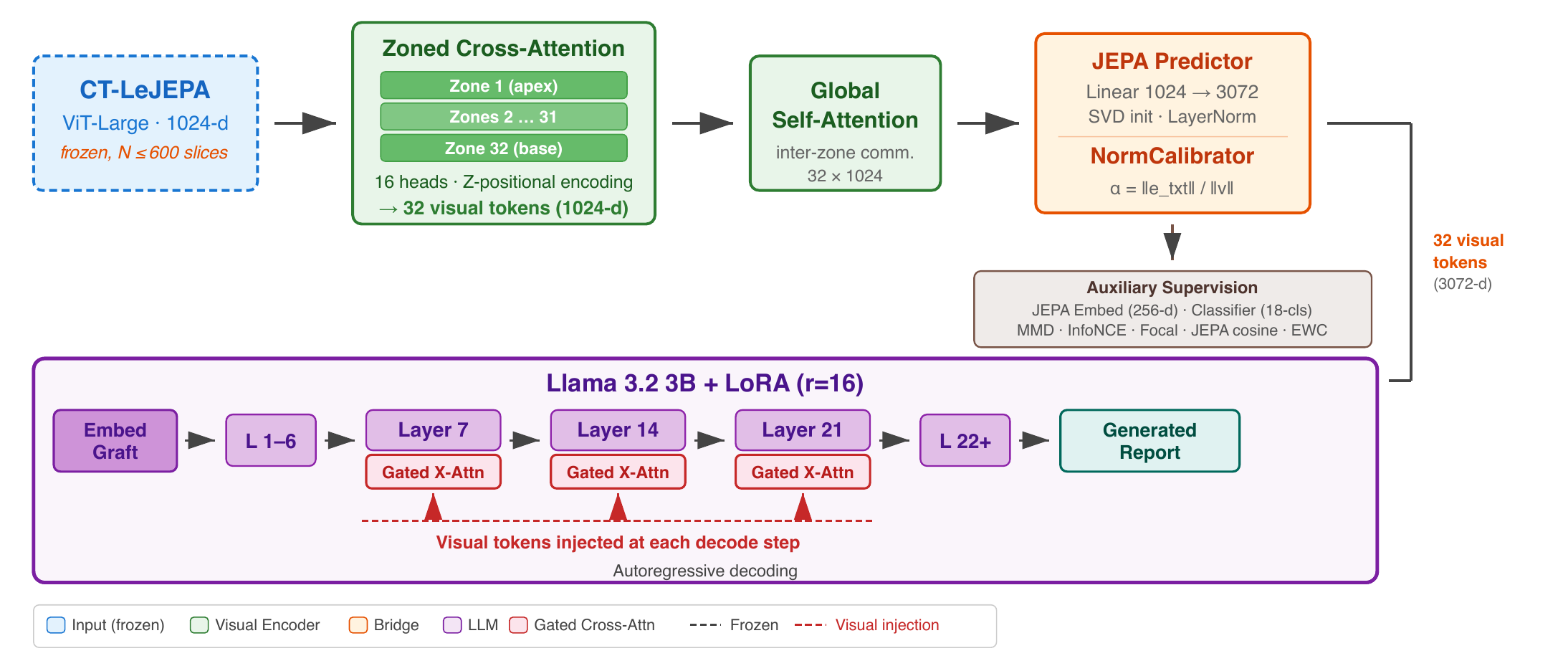}
    \caption{Overview of the Ker-VLJEPA-3B architecture. CT slice embeddings from a frozen LeJEPA ViT-Large are compressed into 32 spatially-grounded tokens via zone-constrained cross-attention, projected to LLM space via the JEPA predictor with norm calibration, and grafted into Llama~3.2~3B at both the embedding level and intermediate layers (7, 14, 21) through gated cross-attention adapters. Auxiliary branches provide supervision via the JEPA embedding head (256-d whitened space) and an 18-class classifier.}
    \label{fig:architecture}
\end{figure*}

Ker-VLJEPA-3B comprises three main components: (1) a visual encoder that compresses variable-length slice embeddings into fixed-size visual tokens, (2) a JEPA predictor that projects visual tokens into the LLM's embedding space, and (3) a Llama~3.2~3B~\cite{llama} decoder with LoRA~\cite{lora} adapters and Flamingo-style gated cross-attention adapters at intermediate layers. Fig.~\ref{fig:architecture} illustrates the complete architecture.

\subsubsection{Visual Encoder: Zone-Constrained Cross-Attention}

Existing approaches to volumetric compression face a fundamental tradeoff between computational cost and spatial fidelity. 3D spatial pooling perceivers~\cite{m3dlamed} reconstruct tokens into 3D coordinates before cross-attention compression, preserving topology but producing large token counts that strain LLM context windows. Dual-stream MLP-mixers~\cite{med3dvlm} blend multi-scale features but rely on computationally expensive decomposed 3D convolutions in the base encoder and provide no explicit spatial grounding in the output tokens. Wavelet-based approaches~\cite{btb3d} operate in the frequency domain but require causal convolution architectures. Text-guided inter-slice scoring~\cite{med2e3} achieves effective filtering but discards slices entirely and requires a paired text query at inference. Most critically, all of these methods either resize volumes to a fixed 3D grid (discarding native resolution) or apply heuristic slice selection that risks eliminating diagnostically relevant slices.

We introduce \textit{zone-constrained cross-attention}, a compression mechanism that processes \textit{all} input slices at native resolution without filtering, resizing, or specialized 3D encoders. The mechanism enforces an inductive bias reflecting the spatial structure of CT volumes: the z-axis is partitioned into 32 contiguous anatomical zones, and each learnable region query attends exclusively to slices within its zone, producing exactly 32 spatially-grounded visual tokens regardless of the input sequence length ($N \leq 600$). This yields an aggressive compression ratio (up to $\sim$19:1) while preserving anatomical localization by construction: token~0 always corresponds to the thoracic apex and token~31 to the base, a property no competing projector architecture guarantees.

The input slice embeddings are first augmented with physical z-positional encoding derived from DICOM z-spacing metadata:
\begin{equation}
    \mathbf{s}_i' = \mathbf{s}_i + \text{PE}(z_i)
\end{equation}
where $\text{PE}(\cdot)$ denotes sinusoidal positional encoding computed from the physical z-coordinate of slice $i$.

We define $K = 32$ learnable region queries $\{\mathbf{q}_k\}_{k=1}^{K}$, initialized via SVD of actual slice embeddings from the training set. The z-axis is partitioned into $K$ contiguous zones, with boundaries computed dynamically based on the number of valid slices $N$ in each volume:
\begin{equation}
    \mathcal{Z}_k = \left\{ i : \left\lfloor \frac{(k-1) \cdot N}{K} \right\rfloor \leq i < \left\lfloor \frac{k \cdot N}{K} \right\rfloor \right\}
\end{equation}

Each region query $\mathbf{q}_k$ attends exclusively to slices within its zone $\mathcal{Z}_k$ via multi-head cross-attention ($H = 16$ heads, $d = 1024$):
\begin{equation}
    \mathbf{v}_k = \text{MHA}(\mathbf{q}_k, \{\mathbf{s}_i'\}_{i \in \mathcal{Z}_k}, \{\mathbf{s}_i'\}_{i \in \mathcal{Z}_k})
\end{equation}
yielding $K = 32$ visual tokens $\mathbf{V} = \{\mathbf{v}_1, \ldots, \mathbf{v}_{32}\} \in \mathbb{R}^{32 \times 1024}$.

A subsequent global self-attention layer (TransformerEncoderLayer, 16 heads) enables inter-zone communication:
\begin{equation}
    \mathbf{V}' = \text{TransformerEncoder}(\mathbf{V})
\end{equation}

\subsubsection{JEPA Predictor and Norm Calibration}

The visual tokens are projected from the visual encoder's representation space ($d_v = 1024$) to the LLM's hidden dimension ($d_{\ell} = 3072$) via the JEPA predictor:
\begin{equation}
    \hat{\mathbf{V}} = \text{LN}(\text{Linear}(\text{Dropout}(\mathbf{V}')))
\end{equation}
where the linear layer is initialized via SVD of text embedding principal components to provide a favorable starting geometry.

A critical implementation detail is \textit{norm calibration}. Pre-trained LLM text embeddings have a characteristic norm distribution (measured mean norm\,=\,1.1484 for Llama~3.2~3B). Visual tokens with significantly different norms are treated as anomalous inputs by the LLM's attention mechanism, degrading performance. The NormCalibrator applies a learned scalar:
\begin{equation}
    \tilde{\mathbf{V}} = \alpha \cdot \hat{\mathbf{V}}, \quad \alpha = \frac{\|\mathbf{e}_{\text{text}}\|}{\|\hat{\mathbf{V}}\|}
\end{equation}
where $\alpha$ is recalibrated at initialization and periodically during training.

\subsubsection{Embedding Grafting and Multi-Layer Injection}

The 32 norm-calibrated visual tokens are grafted into the LLM's input via a chat template containing 32 \texttt{<|visual\_region|>} placeholder tokens. At the embedding level, placeholder token embeddings are replaced with $\tilde{\mathbf{V}}$ via differentiable mask-based scattering.

Beyond input-level grafting, visual information is injected at LLM layers 7, 14, and 21 via \textit{gated cross-attention adapters} following the Flamingo architecture~\cite{flamingo}. At each injection layer $l \in \{7, 14, 21\}$:
\begin{equation}
    \mathbf{h}_l' = \mathbf{h}_l + \text{MHA}_l^{\text{xattn}}(\mathbf{h}_l, \text{Linear}_l(\tilde{\mathbf{V}}), \text{Linear}_l(\tilde{\mathbf{V}}))
\end{equation}
where $\text{Linear}_l$ is a per-layer projector mapping visual tokens to layer $l$'s representation space, and $\text{MHA}_l^{\text{xattn}}$ contains learned Q/K/V/O projections with Xavier initialization (gain\,=\,0.3).

\textbf{Critical implementation detail}: During autoregressive generation, the cross-attention hooks must fire on \textit{every} decode step, not just during the prefill pass. We identified and corrected a bug where a sequence-length guard caused cross-attention to be skipped during token-by-token generation, resulting in a 2.5$\times$ generation F1 improvement (0.122\,$\rightarrow$\,0.304).

\subsubsection{JEPA Embedding Head and PCA Whitening}
\label{sec:whitening}

To provide non-autoregressive semantic supervision, we train a JEPA embedding head that projects pooled visual tokens into a whitened text embedding space:
\begin{equation}
    \mathbf{z}_v = \text{Linear}_2(\text{GELU}(\text{LN}(\text{Linear}_1(\bar{\mathbf{V}}))))
\end{equation}
where $\bar{\mathbf{V}} = \text{mean}(\tilde{\mathbf{V}}) \in \mathbb{R}^{3072}$ and $\mathbf{z}_v \in \mathbb{R}^{256}$.

Raw LLM text embeddings suffer from severe anisotropy, a well-documented phenomenon where embeddings cluster in a narrow cone with nearly uniform pairwise similarity. We measured catastrophic anisotropy in Llama~3.2~3B layer~14 representations of CT-RATE reports (Table~\ref{tab:whitening}), rendering standard contrastive learning (InfoNCE) ineffective due to a nearly flat loss surface.

\begin{table}[t]
    \centering
    \caption{Effect of PCA Whitening on Text Embedding Quality}
    \label{tab:whitening}
    \begin{tabular}{lcc}
        \toprule
        Metric & Before (3072d) & After (256d) \\
        \midrule
        Mean pairwise cosine sim. & 0.949 & $-$0.001 \\
        Discriminability ($d'$) & 1.36 & 16.03 \\
        Variance retained & --- & 97.3\% \\
        \bottomrule
    \end{tabular}
\end{table}

We resolve this by applying PCA whitening: the top 256 principal components of 22{,}773 training report embeddings are used to project from the anisotropic 3072-d space to an isotropic 256-d space, achieving an 11.8$\times$ improvement in discriminability ($d' = 1.36 \rightarrow 16.03$) while retaining 97.3\% of variance.

\subsection{Four-Phase Curriculum Learning}

\begin{figure*}[t]
    \centering
    \includegraphics[width=0.95\textwidth]{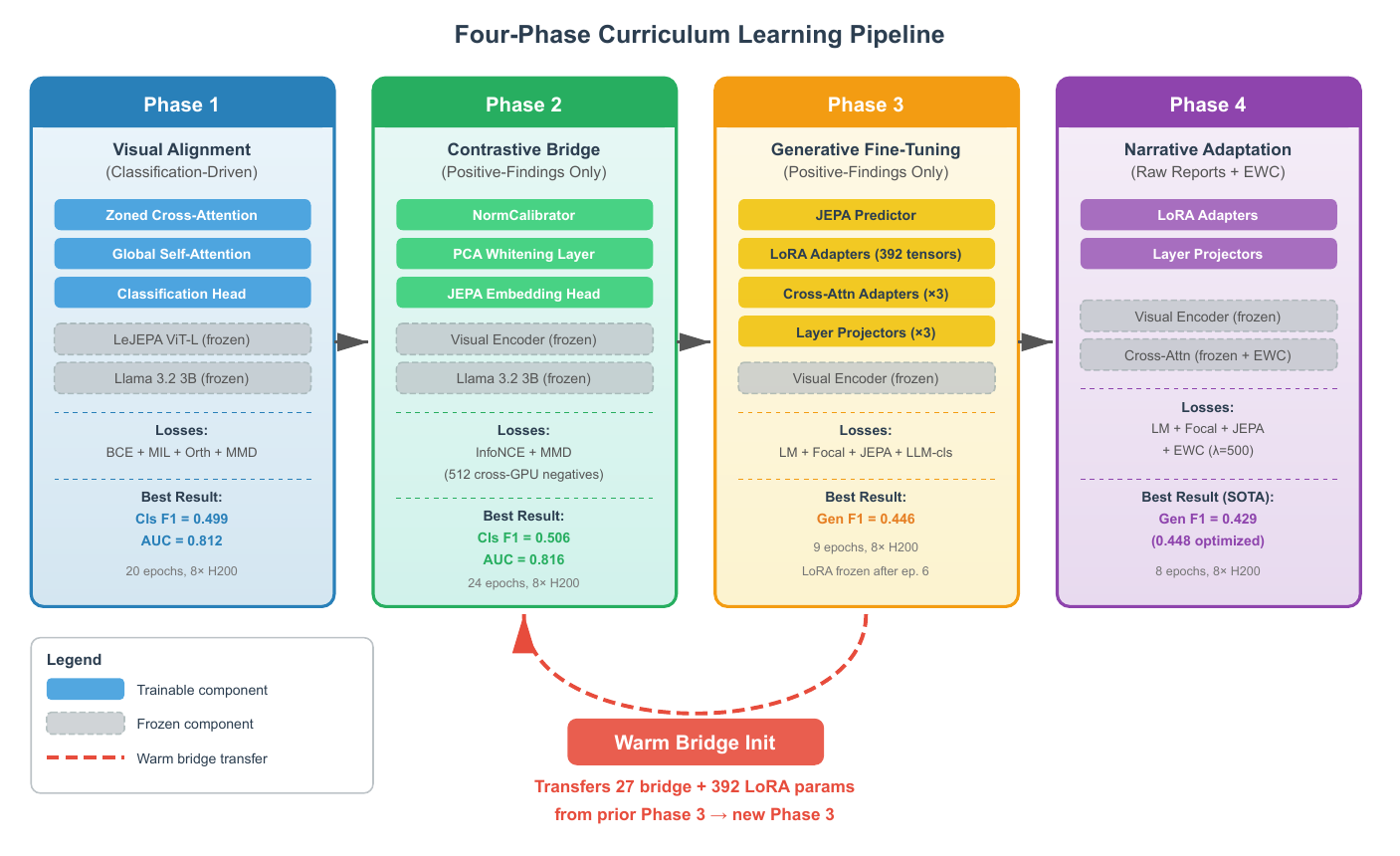}
    \caption{Four-phase curriculum learning pipeline. Each phase progressively builds capability from visual discrimination (Phase~1) through contrastive alignment (Phase~2) to free-text generation (Phase~3) and narrative adaptation (Phase~4). Colored boxes indicate trainable components; dashed boxes indicate frozen components. The warm bridge initialization (red dashed arrow) transfers 27 converged bridge components and 392 LoRA tensors from a prior Phase~3 run, providing immediate convergence and eliminating the cold-start problem.}
    \label{fig:curriculum}
\end{figure*}

The training pipeline follows a four-phase curriculum that progressively builds capability from visual discrimination to free-text generation (Fig.~\ref{fig:curriculum}).

\subsubsection{Phase 1: Visual Alignment (Classification-Driven)}

Phase~1 trains the visual encoder to produce discriminative tokens for 18-class abnormality detection. The LLM and JEPA predictor are frozen. The combined loss is:
\begin{equation}
    \mathcal{L}_1 = \lambda_{\text{cls}} \mathcal{L}_{\text{BCE}} + \lambda_{\text{mil}} \mathcal{L}_{\text{MIL}} + \lambda_{\text{orth}} \mathcal{L}_{\text{orth}} + \lambda_{\text{mmd}} \mathcal{L}_{\text{MMD}}
\end{equation}
with weights $\lambda_{\text{cls}}=1.5$, $\lambda_{\text{mil}}=1.0$, $\lambda_{\text{orth}}=1.0$, $\lambda_{\text{mmd}}=0.5$.

The classification loss uses binary cross-entropy with per-class positive weights ($w_c = \min(n_{\text{neg}}/n_{\text{pos}}, 10)$). The MIL (multiple instance learning) loss applies max-pooling over region tokens before classification, providing a secondary learning signal. The orthogonality loss encourages diverse visual token representations.

\textbf{Per-condition MMD alignment.} Following the KerJEPA framework~\cite{kerjepa}, which established that kernel-based discrepancy regularizers yield provable gains in training stability for joint-embedding architectures, we introduce a distributional alignment loss using Maximum Mean Discrepancy (MMD)~\cite{mmd} with an inverse multi-quadratic (IMQ) kernel in the 256-d whitened space. For each sample, the 32 visual tokens are matched against that sample's $K$ positive-condition text embeddings:
\begin{equation}
    \mathcal{L}_{\text{MMD}} = \mathbb{E}[k(\mathbf{z}_v, \mathbf{z}_v')] - 2\mathbb{E}[k(\mathbf{z}_v, \mathbf{z}_t)] + \mathbb{E}[k(\mathbf{z}_t, \mathbf{z}_t')]
\end{equation}
where $k(\mathbf{x}, \mathbf{y}) = (1 + \alpha\|\mathbf{x} - \mathbf{y}\|^2)^{-1/2}$ with $\alpha = 4\gamma/(2D-3) \approx 0.039$ for $\gamma=5.0$, $D=256$. Normal volumes ($K=0$) receive no MMD loss.

\subsubsection{Phase 2: Contrastive Bridge Training}

Phase~2 aligns visual and text representations via InfoNCE contrastive learning with cross-GPU negative mining (512 effective negatives from 8 GPUs) and a learned temperature parameter (CLIP-style, initialized at $\tau = 0.10$):
\begin{equation}
    \mathcal{L}_{\text{NCE}} = -\log \frac{\exp(\text{sim}(\mathbf{z}_v^i, \mathbf{z}_t^i)/\tau)}{\sum_{j=1}^{B} \exp(\text{sim}(\mathbf{z}_v^i, \mathbf{z}_t^j)/\tau)}
\end{equation}

Critically, Phase~2 uses \textit{positive-findings-only text} for contrastive targets. Aligning visual tokens against raw reports (90\% normal text) would train the visual encoder to map all scans toward a uniform ``normal'' embedding. Using only positive-condition text descriptions ensures that visual-text alignment captures pathology-discriminative information.

\subsubsection{Phase 3: Generative Fine-Tuning}
\label{sec:phase3}

Phase~3 is the most critical phase, training the model to generate free-text reports. The visual encoder is frozen; trainable parameters include the JEPA predictor, LoRA adapters, cross-attention adapters, and layer projectors.

\textbf{The posterior collapse problem.} In all early experimental runs, the generation F1 peaked at epochs~1--4 and then collapsed to $\sim$0.15 as the LLM converged to generating fluent, generic normal-anatomy descriptions using language priors alone (Table~\ref{tab:collapse}). The root cause is a \textit{structural gradient imbalance}: approximately 90\% of tokens in raw radiology reports describe normal anatomy (``Trachea and both main bronchi are open...''), overwhelming the 10\% that describe pathological findings. The language modeling loss gradient is thus dominated by normal-text tokens, teaching the LLM to ignore visual features.

\textbf{Solution: positive-findings-only training.} Instead of raw narrative reports, the training text consists of per-class natural language narrative segments for only the positive conditions present in each scan. This eliminates normal-text tokens entirely, so every token the model trains on is clinically relevant to a finding the visual encoder detected. The order of conditions is randomized per sample to prevent memorizing a fixed sequence. Normal volumes ($\sim$35--40\% of data) receive a short template: ``No significant thoracic abnormalities identified.''

\begin{table}[t]
    \centering
    \caption{Phase 3 Posterior Collapse: Historical Experimental Comparison}
    \label{tab:collapse}
    \begin{tabular}{lccl}
        \toprule
        Run & Best F1 & Epochs to & Outcome \\
         & & collapse & \\
        \midrule
        Baseline & 0.198 & 2 & Never sustained \\
        + Cross-attn fix & 0.304 & 1 & Immediate collapse \\
        + Vis. dropout & 0.262 & 1 & Faster collapse \\
        + LLM visual cls. & 0.259 & 4 & Slight delay \\
        \midrule
        \textbf{+ Pos.-findings} & \textbf{0.427} & \textbf{none} & \textbf{Sustained $>$0.40} \\
        \textbf{+ Warm bridge} & \textbf{0.446} & \textbf{none} & \textbf{New SOTA} \\
        \bottomrule
    \end{tabular}
\end{table}

The Phase~3 loss combines language modeling, focal classification, and JEPA embedding prediction:
\begin{equation}
    \mathcal{L}_3 = \mathcal{L}_{\text{LM}} + \lambda_{\text{fcls}} \mathcal{L}_{\text{focal}} + \lambda_{\text{jepa}} \mathcal{L}_{\text{JEPA}} + \lambda_{\text{vcls}} \mathcal{L}_{\text{LLM-cls}}
\end{equation}
where $\mathcal{L}_{\text{focal}} = -(1 - p_t)^\gamma \log(p_t)$~\cite{focal} with $\gamma = 2.0$, $\lambda_{\text{vcls}} = 3.0$, and the LLM visual classification loss operates on last-layer hidden states to force visual information preservation through all LLM layers. Label masking ensures loss is computed only on assistant response tokens.

Additional Phase~3 training techniques include: LoRA freezing after epoch~6 (preventing language-prior shortcuts), ReduceLROnPlateau on gen\_f1 (patience\,=\,3, factor\,=\,0.5), and adaptive projector learning rate scaling (LARS-inspired, 1--30$\times$ range).

\textbf{Warm bridge initialization.} A fundamental problem in curriculum-based multimodal training is that transitioning between phases resets unmapped parameters. In Phase~3, the 27 bridge components (3~layer projectors, 21~cross-attention adapter parameters) and 392~LoRA tensors are absent from the Phase~2 checkpoint and initialize randomly, regardless of Phase~2 quality. We verified this empirically: improving Phase~2 with InfoNCE\,+\,MMD but using a cold bridge yielded F1\,=\,0.424, slightly worse than the baseline 0.427.

The warm bridge technique loads converged bridge components from a prior Phase~3 run:
\begin{enumerate}
    \item 3 layer projectors (visual$\rightarrow$LLM linear maps)
    \item 3 cross-attention adapters (Q/K/V/O projections + layer norms, 21 parameter tensors)
    \item 392 LoRA weight tensors
\end{enumerate}
After loading Phase~2 weights, the warm-init code selectively overwrites only these 416 bridge components. The visual tokens from the improved Phase~2 are in a sufficiently similar distribution (same architecture, same training objectives) that the converged bridge transfers effectively.

\subsubsection{Phase 4: Raw Narrative Fine-Tuning}

Phases~2--3 train on positive-findings-only text to avoid gradient domination. However, the CT-RATE evaluation expects full narrative reports including negative findings. Phase~4 fine-tunes the model on raw radiologist narrative text (verbatim Findings\_EN from CT-RATE) to produce authentic clinical prose while preserving pathology detection.

\textbf{The catastrophic forgetting challenge.} Five experimental iterations were required to find a working configuration (Table~\ref{tab:phase4_history}). The key insight is that cross-attention adapters govern \textit{visual grounding} (what the model attends to), while LoRA governs \textit{generative style} (how the model generates text). When Phase~4 trains on raw narrative text (90\% normal anatomy), the language modeling gradient corrupts cross-attention attention patterns, destroying pathology detection.

\begin{table}[t]
    \centering
    \caption{Phase 4 Experimental History: Discovering the Correct Configuration}
    \label{tab:phase4_history}
    \begin{tabular}{clc}
        \toprule
        Ver. & Configuration & Best F1 \\
        \midrule
        v1 & No freeze, no EWC & 0.297$\downarrow$ \\
        v2 & Freeze LoRA (3 epochs) & 0.289$\downarrow$ \\
        v3 & EWC + JEPA + MMD & 0.274 \\
        v4 & EWC only & 0.304$\downarrow$ \\
        \textbf{v5} & \textbf{Freeze cross-attn + EWC} & \textbf{0.418} \\
        \bottomrule
    \end{tabular}
\end{table}

\textbf{Solution: selective freezing with EWC.} Cross-attention adapters and layer projectors are frozen ($\texttt{requires\_grad=False}$). Only LoRA adapters are trainable, constrained by EWC:
\begin{equation}
    \mathcal{L}_4 = \mathcal{L}_{\text{LM}} + \lambda_{\text{cls}} \mathcal{L}_{\text{focal}} + \lambda_{\text{ewc}} \sum_{i} (\theta_i^{\text{LoRA}} - \theta_i^{\text{P3}})^2
\end{equation}
where $\theta_i^{\text{P3}}$ are Phase~3 LoRA reference weights captured at training start, and $\lambda_{\text{ewc}} = 100$. The ultra-conservative learning rate ($5 \times 10^{-7}$, 40$\times$ lower than Phase~3) provides additional stability.

\section{Experiments}
\label{sec:experiments}

\subsection{Dataset and Evaluation Protocol}

\textbf{CT-RATE benchmark.} We evaluate on CT-RATE~\cite{ctrate}, comprising $\sim$46{,}400 training and $\sim$3{,}000 validation thoracic CT volumes with 18 binary abnormality labels and radiologist-authored narrative reports. After filtering for available embeddings, 2{,}984 validation volumes are used for evaluation.

\textbf{Evaluation protocol.} Following the established protocol used by CT-CLIP~\cite{ctclip}, CT-CHAT~\cite{ctchat}, BTB3D~\cite{btb3d}, and U-VLM~\cite{uvlm}: (1)~generate free-text narrative reports from CT volumes (temperature\,=\,0.6, top\_p\,=\,0.9); (2)~extract 18 binary labels using the official CT-RATE RadBERT~\cite{radbert} classifier; (3)~compute macro-averaged F1, precision, and recall against ground-truth labels. The RadBERT classifier achieves macro~F1\,=\,0.982 on ground-truth reports, confirming it is not a bottleneck.

\textbf{Hardware.} All experiments use 8$\times$ NVIDIA H200 GPUs with DDP via HuggingFace Accelerate, bf16 mixed precision.

\subsection{Implementation Details}

\textbf{Visual encoder input.} CT volumes are pre-processed by the Guided-Chest-CT-LeJEPA~\cite{lejepa_paper,lejepa} backbone, a ViT-Large ({\tt vit\_large\_patch14\_dinov2} architecture, trained from scratch) that was self-supervised on the CT-RATE training split using a Latent-Euclidean Joint-Embedding Predictive Architecture with anatomy-guided semi-3D cropping and an auxiliary 118-class organ prediction objective. Each axial slice is encoded into a 1024-dimensional embedding; a volume of up to 600 slices yields the input sequence $\mathbf{S} \in \mathbb{R}^{N \times 1024}$. The LeJEPA weights are frozen throughout all four training phases.

\textbf{LLM backbone.} Llama~3.2~3B Instruct~\cite{llama} serves as the frozen language model backbone. LoRA~\cite{lora} adapters ($r=16$) are applied to all attention layers. Gated cross-attention adapters are injected at layers~7, 14, and 21.

\textbf{Phase-specific hyperparameters.} Phase~1: 20 epochs, batch size 32, LR\,=\,$5 \times 10^{-5}$, cosine decay. Phase~2: 30 max epochs (early stopped at 24, patience\,=\,8), batch size 64/GPU (512 effective), LR\,=\,$3 \times 10^{-5}$. Phase~3: 50 max epochs (best at epoch~9 with warm bridge), batch size 8, LR\,=\,$2 \times 10^{-5}$. Phase~4: 30 max epochs (best at epoch~8), batch size 8, LR\,=\,$5 \times 10^{-7}$.

\textbf{Generation configuration.} Temperature\,=\,0.6 (confirmed best via hyperparameter sweep), top\_p\,=\,0.9, repetition\_penalty\,=\,1.15, no\_repeat\_ngram\_size\,=\,5, max\_new\_tokens\,=\,384.

\subsection{Main Results}

\subsubsection{Comparison with State-of-the-Art}

Table~\ref{tab:sota} presents the comparison with existing methods on the CT-RATE benchmark. All methods are evaluated using the identical protocol: generate reports, extract labels via the official RadBERT classifier, compute macro metrics.

\begin{table}[t]
    \centering
    \caption{Comparison with State-of-the-Art on CT-RATE Benchmark (2{,}984 Validation Volumes)}
    \label{tab:sota}
    \begin{tabular}{lccc}
        \toprule
        Method & Macro F1 & Macro Prec & Macro Rec \\
        \midrule
        CT-CLIP~\cite{ctclip} & 0.194 & --- & --- \\
        CT-CHAT~\cite{ctchat} & 0.287 & --- & --- \\
        BTB3D~\cite{btb3d} & 0.354 & --- & --- \\
        U-VLM~\cite{uvlm} & 0.414 & 0.491 & 0.429 \\
        \midrule
        Ker-VLJEPA-3B (P3) & 0.422 & 0.380 & 0.517 \\
        \textbf{Ker-VLJEPA-3B (P4)} & \textbf{0.429} & \textbf{0.389} & \textbf{0.524} \\
        Ker-VLJEPA-3B (P4, opt.) & 0.448 & 0.380 & 0.585 \\
        \bottomrule
    \end{tabular}
\end{table}

Ker-VLJEPA-3B Phase~4 achieves macro~F1\,=\,0.429, surpassing U-VLM by +3.6\% at the default threshold. The improvement is driven primarily by substantially higher recall (+22.1\%), while U-VLM maintains a precision advantage (+26.2\%). Per-class threshold optimization yields macro~F1\,=\,0.448 (+8.2\% over U-VLM), though this has a data leakage caveat as thresholds are optimized on the evaluation set.

\subsubsection{Per-Phase Progression}

Table~\ref{tab:phases} shows the progression across all four curriculum phases, demonstrating that each phase contributes meaningfully to the final result.

\begin{table}[t]
    \centering
    \caption{Performance Progression Across Curriculum Phases (Full Validation Set)}
    \label{tab:phases}
    \begin{tabular}{llcccc}
        \toprule
        Phase & Type & F1 & Prec & Rec & AUC \\
        \midrule
        1 & Cls. & 0.460 & 0.463 & 0.551 & 0.811 \\
        2 & Cls. & 0.465 & 0.490 & 0.532 & 0.816 \\
        3 & Gen. & 0.422 & 0.380 & 0.517 & --- \\
        \textbf{4} & \textbf{Gen.} & \textbf{0.429} & \textbf{0.389} & \textbf{0.524} & --- \\
        \bottomrule
    \end{tabular}
\end{table}

Phase~1 achieves strong classification performance (F1\,=\,0.460, AUC\,=\,0.811). Phase~2 contrastive training improves both F1 (+1.1\%) and AUC (+0.6\%) across all 18 classes. Phase~3 generation shows a moderate gap between classification and generation metrics (0.465\,$\rightarrow$\,0.422), reflecting the inherent difficulty of converting discriminative visual features into accurate free-text reports. Critically, Phase~4 \textit{improves} over Phase~3 (0.422\,$\rightarrow$\,0.429)---a key achievement enabled by our selective freezing strategy, as V1 (without cross-attention freezing) showed Phase~4 regression (0.423\,$\rightarrow$\,0.399).

\subsubsection{Warm Bridge Effectiveness}

Table~\ref{tab:warm_bridge} demonstrates the dramatic impact of warm bridge initialization on Phase~3 training dynamics.

\begin{table}[t]
    \centering
    \caption{Cold vs.\ Warm Bridge Phase~3 Training Comparison}
    \label{tab:warm_bridge}
    \begin{tabular}{lccc}
        \toprule
        Metric & Cold & Cold + & \textbf{Warm} \\
        & Baseline & Better P2 & \textbf{Bridge} \\
        \midrule
        Best F1 & 0.427 (ep16) & 0.424 (ep7) & \textbf{0.446 (ep9)} \\
        Best Precision & 0.391 & 0.366 & \textbf{0.437} \\
        Epoch 1 F1 & 0.360 & 0.362 & \textbf{0.425} \\
        Epochs to $>$0.42 & 5 & 5 & \textbf{1} \\
        Val.\ loss floor & 1.130 & 1.069 & \textbf{1.056} \\
        \bottomrule
    \end{tabular}
\end{table}

The warm bridge provides immediate convergence: epoch~1 F1\,=\,0.425 versus 0.360 for cold start (+18\%). Importantly, improving Phase~2 quality alone (cold bridge + better P2) does not help (F1\,=\,0.424), demonstrating that the bridge reset dominates over representation quality. The warm bridge also yields substantially better precision (0.437 vs.\ 0.391), generating fewer false-positive findings.

\subsubsection{Per-Class Results}

Table~\ref{tab:perclass} reports per-class generation results for the final model (Phase~4, full validation set). Performance varies substantially across conditions, with F1 ranging from 0.664 (pleural effusion) to 0.203 (interlobular septal thickening), correlating with class prevalence and the distinctiveness of imaging findings.

\begin{table}[t]
    \centering
    \caption{Per-Class Results (Phase~4, 2{,}984 Validation Volumes)}
    \label{tab:perclass}
    \begin{tabular}{lcccr}
        \toprule
        Class & Prec & Rec & F1 & Support \\
        \midrule
        Pleural effusion & 0.574 & 0.789 & 0.664 & 370 \\
        Art.\ wall calcification & 0.642 & 0.660 & 0.651 & 849 \\
        Coronary artery calc. & 0.580 & 0.585 & 0.582 & 752 \\
        Cardiomegaly & 0.437 & 0.743 & 0.550 & 315 \\
        Lung nodule & 0.491 & 0.560 & 0.523 & 1344 \\
        Lung opacity & 0.545 & 0.453 & 0.494 & 1173 \\
        Emphysema & 0.346 & 0.667 & 0.456 & 588 \\
        Lymphadenopathy & 0.393 & 0.532 & 0.452 & 769 \\
        Pulm.\ fibrotic sequela & 0.384 & 0.505 & 0.436 & 819 \\
        Atelectasis & 0.369 & 0.496 & 0.423 & 698 \\
        Consolidation & 0.430 & 0.389 & 0.408 & 576 \\
        Pericardial effusion & 0.238 & 0.614 & 0.343 & 215 \\
        Mosaic atten.\ pattern & 0.256 & 0.494 & 0.337 & 245 \\
        Hiatal hernia & 0.257 & 0.397 & 0.312 & 413 \\
        Peribronchial thick. & 0.271 & 0.354 & 0.307 & 347 \\
        Medical material & 0.194 & 0.722 & 0.305 & 306 \\
        Bronchiectasis & 0.238 & 0.328 & 0.276 & 326 \\
        Interlob.\ septal thick. & 0.357 & 0.142 & 0.203 & 246 \\
        \midrule
        \textbf{Macro average} & \textbf{0.389} & \textbf{0.524} & \textbf{0.429} & \\
        \bottomrule
    \end{tabular}
\end{table}

\section{Ablation Studies}
\label{sec:ablation}

We conduct a comprehensive three-part visual token ablation study to verify that the model achieves genuine visual grounding, rather than generating reports from language priors alone.

\subsection{Linear Probe: Information Preservation}

Linear probe classification (5-fold CV logistic regression on mean-pooled visual tokens, 2{,}984 samples) measures how much pathology information is preserved across the pipeline (Table~\ref{tab:probe}).

\begin{table}[t]
    \centering
    \caption{Linear Probe F1 Across Pipeline Stages}
    \label{tab:probe}
    \begin{tabular}{lccc}
        \toprule
        Stage & Dim & F1 (macro) \\
        \midrule
        Raw LeJEPA embeddings & 1024 & 0.447 \\
        Phase 1 (visual context) & 1024 & 0.459 \\
        Phase 1 (norm-matched) & 3072 & 0.488 \\
        Phase 2 (norm-matched) & 3072 & \textbf{0.495} \\
        Phase 3 (norm-matched) & 3072 & 0.495 \\
        Phase 4 (norm-matched) & 3072 & 0.495 \\
        \bottomrule
    \end{tabular}
\end{table}

Key findings: (1)~the visual encoder adds +10.7\% discriminative information beyond raw slice features (0.447\,$\rightarrow$\,0.495); (2)~Phase~2 contrastive training improves representations (+1.4\%); (3)~freezing the visual encoder in Phases~3--4 perfectly preserves probe F1 (0.495 in Phase~2, 3, and 4), validating the frozen encoder design.

\subsection{Generation Ablation: Visual Token Contribution}

Generation F1 under four visual token conditions (304 samples, Table~\ref{tab:gen_ablation}):

\begin{table}[t]
    \centering
    \caption{Generation Ablation: F1 Under Different Visual Token Conditions}
    \label{tab:gen_ablation}
    \begin{tabular}{lcccc}
        \toprule
        Condition & P3 F1 & P3 Prec & P4 F1 & P4 Prec \\
        \midrule
        \textbf{Normal} & \textbf{0.222} & 0.408 & \textbf{0.190} & 0.298 \\
        Zeroed & 0.096 & 0.114 & 0.104 & 0.146 \\
        Random & 0.121 & 0.156 & 0.121 & 0.129 \\
        Shuffled & 0.122 & 0.204 & 0.119 & 0.137 \\
        \bottomrule
    \end{tabular}
\end{table}

Zeroing visual tokens destroys 56.6\% (Phase~3) and 44.9\% (Phase~4) of generation F1. Crucially, \textit{shuffled} tokens (from a different patient) perform no better than random noise, proving the LLM reads patient-specific pathology content from the visual tokens, not merely their statistical properties. Precision collapses from 0.408 to 0.114 without visual tokens, indicating the model over-generates findings indiscriminately when it cannot ``see'' what is present.

\subsection{NLL Ablation: Semantic Binding}

Teacher-forced negative log-likelihood analysis (200 samples, Table~\ref{tab:nll}) reveals that visual tokens provide 2$\times$ stronger contribution on pathology-specific words ($\Delta$NLL\,=\,+0.020) compared to generic text ($\Delta$NLL\,=\,+0.011), confirming that cross-attention adapters inject pathology information at semantically appropriate positions.

\begin{table}[t]
    \centering
    \caption{NLL Ablation: Impact of Visual Token Manipulation}
    \label{tab:nll}
    \begin{tabular}{lcccc}
        \toprule
        Condition & P3 NLL & $\Delta$ & P4 NLL & $\Delta$ \\
        \midrule
        Correct & 1.200 & --- & 1.196 & --- \\
        Zeroed & 1.211 & +0.9\% & 1.209 & +1.1\% \\
        Shuffled & 1.264 & +5.3\% & 1.259 & +5.3\% \\
        \bottomrule
    \end{tabular}
\end{table}

Notably, shuffled tokens cause a 5.3\% NLL increase, substantially larger than the 0.9--1.1\% increase from zeroed tokens. This is because zeroed tokens represent a ``known unknown'' (the model learns cautious generation), while wrong tokens actively mislead the model.

\subsection{Component Ablation}

Table~\ref{tab:component_ablation} summarizes the contribution of each key methodological innovation.

\begin{table}[t]
    \centering
    \caption{Component Ablation: Impact of Key Innovations}
    \label{tab:component_ablation}
    \begin{tabular}{lcc}
        \toprule
        Configuration & F1 & $\Delta$ vs.\ Base \\
        \midrule
        Cross-attn fix only & 0.304 & --- \\
        + Positive-findings-only & 0.427 & +0.123 \\
        + Warm bridge & 0.446 & +0.142 \\
        + Phase 4 (freeze xattn + EWC) & 0.429$^*$ & +0.125 \\
        \bottomrule
        \multicolumn{3}{l}{\footnotesize $^*$Phase~4 generates raw narrative; direct comparison with} \\
        \multicolumn{3}{l}{\footnotesize positive-findings Phase~3 reflects task difference, not regression.}
    \end{tabular}
\end{table}

The positive-findings-only strategy provides the largest single improvement (+0.123 F1), followed by warm bridge initialization (+0.019 over cold-start positive-findings). The cross-attention generation fix was foundational, providing 2.5$\times$ improvement from the pre-fix baseline.

\section{Discussion}
\label{sec:discussion}

\subsection{Recall-Precision Trade-off}

Ker-VLJEPA-3B achieves substantially higher recall than U-VLM (+22.1\%) at the cost of lower precision ($-$20.8\%). This reflects a design philosophy prioritizing sensitivity: in clinical settings, failing to mention a finding (false negative) is generally more consequential than mentioning one that is equivocal (false positive), as radiologists can readily dismiss over-reported findings but cannot evaluate omitted ones. The warm bridge substantially improved precision over the V1 baseline (0.389 vs.\ 0.329), indicating this gap is narrowing. Future work could explore precision-oriented training signals, calibrated confidence thresholds, or ensemble methods.

\subsection{Why Positive-Findings-Only Training Works}

The positive-findings-only strategy addresses a fundamental architectural asymmetry in vision-language models for radiology. When training on raw reports, the LLM receives two sources of signal for generating normal-anatomy text: (1)~language priors from pre-training, and (2)~the visual tokens. For pathological findings, only the visual tokens provide the relevant signal. Since language priors are overwhelmingly strong (Llama~3.2~3B was pre-trained on trillions of tokens), the gradient from normal-text tokens reinforces language priors rather than visual attention. By removing normal-text tokens, every gradient update necessarily engages the visual pathway, preventing the model from learning a visual-token-ignoring shortcut.

\subsection{The Bridge Reset Problem}

Our finding that improving Phase~2 representation quality does not translate to Phase~3 generation quality (when using cold bridge initialization) has broad implications for curriculum-based multimodal training. The 416 randomly-initialized bridge parameters (projectors, cross-attention adapters, LoRA) at Phase~3 start represent a bottleneck that is independent of upstream representation quality. The warm bridge technique resolves this by decoupling representation learning from bridge training, allowing each to be optimized independently and then composed.

\subsection{Limitations}

Several limitations should be noted. First, evaluation is performed on a single benchmark (CT-RATE); multi-center validation across different scanner manufacturers and clinical populations would strengthen generalizability claims. Second, the pre-computed LeJEPA slice embeddings are treated as fixed inputs; end-to-end training from raw voxels may capture finer-grained features. Third, the per-class threshold optimization (F1\,=\,0.448) involves data leakage and should be interpreted as an upper bound. Fourth, the current model requires 8$\times$ H200 GPUs for training, limiting accessibility. Fifth, we do not include a formal observer study with clinical radiologists, which would strengthen the clinical validation.

\subsection{Clinical Implications}

While macro~F1\,=\,0.429 is not sufficient for autonomous report generation, it demonstrates meaningful progress toward clinically useful decision support. The model's high recall profile makes it particularly suited as a ``safety net'' that flags potential findings for radiologist review. The zone-constrained architecture provides implicit spatial grounding (token~0 maps to the thoracic apex, token~31 to the base), potentially enabling future localization of findings.

\section{Conclusion}
\label{sec:conclusion}

We presented Ker-VLJEPA-3B, a four-phase curriculum learning framework for automated 3D CT report generation that achieves state-of-the-art performance on the CT-RATE benchmark (macro~F1\,=\,0.429, +3.6\% over U-VLM). A defining characteristic of our approach is the complete separation of visual representation learning from language: the LeJEPA backbone is trained via purely self-supervised joint-embedding prediction on unlabeled CT volumes, with no text supervision whatsoever. This stands in contrast to all prior work on this benchmark, which relies on vision encoders shaped by text, whether through contrastive image-text pre-training or semantically-derived segmentation labels. Our results demonstrate that language-free visual representations, when connected to a language model through a well-designed curriculum bridge, not only match but surpass text-supervised alternatives. This decoupled design is inherently modality-agnostic: because the bridge and curriculum impose no assumptions about the input encoder's training objective, the same framework can integrate any self-supervised foundation model, whether trained on medical imaging, genomic sequences, audio, or sensor data, into a language model for narrative generation. Our further contributions, including zone-constrained cross-attention, PCA whitening, positive-findings-only training, warm bridge initialization, and selective freezing with EWC, address fundamental challenges in grafting non-linguistic representations into pre-trained language models. The comprehensive ablation study confirms genuine visual grounding with 56.6\% of generation quality deriving from patient-specific visual content. We believe this language-free, modality-agnostic paradigm opens a path toward multimodal AI systems that can leverage the rapidly growing landscape of self-supervised foundation models across domains without requiring paired text for each new modality.


\end{document}